%% file: main.tex
\documentclass[sigconf]{acmart}
\usepackage[moderate]{savetrees}
\usepackage{booktabs} % For formal tables
\usepackage{listings}
\usepackage{latexsym}
\usepackage[sets]{cryptocode}
\usepackage{amsmath}
\usepackage{amssymb}
\usepackage{graphicx}
\usepackage{setspace}
\usepackage{fullpage}
\usepackage{xspace}
\usepackage{xcolor}
\usepackage{caption}
\usepackage{subfigure}
\usepackage{courier}
\usepackage{enumitem}
\usepackage[font=normal,skip=2pt]{caption}
\usepackage{times}
\usepackage{microtype}
\usepackage{balance} % to better equalize the last page
\usepackage{xcolor}
\usepackage[hang,flushmargin]{footmisc}
\renewcommand\footnotetextcopyrightpermission[1]{}
\DeclareMathOperator*{\argmax}{arg\,max}

\setcopyright{none}
\acmConference[]{}{}

\definecolor{tomato}{rgb}{1,0.2,0}
\definecolor{turqoise}{rgb}{0.03, 0.91, 0.87}
\definecolor{grey}{rgb}{0.4,0.4,0.4}
\newif\ifnotes
\notestrue

\DeclareRobustCommand{\subhead}[1]{\noindent\textbf{#1}}
\DeclareRobustCommand{\system}{\mbox{\sc Ruler}\xspace}

\DeclareRobustCommand{\tagruler}{\mbox{\sc TagRuler}\xspace}

\newcommand{\eat}[1]{}
\newcommand{\example}[1]{{\vspace*{3pt}\noindent\underline{Example:} #1\qed}}
\newcommand{\stitle}[1]{\smallskip \noindent{\textbf{#1}}}

\DeclareRobustCommand{\subhead}[1]{\noindent\textbf{#1}} 
\newcommand{\squishlist}{
   \begin{list}{$\bullet$}
    {
      \setlength{\itemsep}{0pt}
      \setlength{\parsep}{3pt}
      \setlength{\topsep}{3pt}
      \setlength{\partopsep}{0pt}
      \setlength{\leftmargin}{1.5em}
      \setlength{\labelwidth}{1em}
      \setlength{\labelsep}{0.5em} } }

\newcommand{\squishend}{ \end{list}  }
    
\renewcommand{\shortauthors}{}
\begin{document}

\title[]{Data Programming by Demonstration:\\A Framework for Interactively Learning Labeling Functions}

\author{Sara Evensen}
\affiliation{%
  \institution{Megagon Labs\vspace{-5pt}}
}
% \email{}
\author{Chang Ge}
\authornote{Work done during internship at Megagon Labs.}
\affiliation{%
  \institution{University of Waterloo\vspace{-5pt}}
}
% \email{}
\author{Dongjin Choi}
\authornotemark[1]
\affiliation{
  \institution{Georgia Tech}
}
% \email{}
\author{\c{C}a\u{g}atay Demiralp}
\affiliation{
  \institution{Megagon Labs}
}
% \email{}

% \renewcommand{\shortauthors}{B. Trovato et al.}

% \renewcommand{\shortauthors}{Evensen et al.}

\input{abstract}

% \keywords{}

\maketitle

\pagestyle{plain}

\input{intro}

\input{framework}

\input{ruler}

\input{evaluation}
\input{results}

\input{related}

\input{tagruler}

\input{discussion}
\bibliographystyle{ACM-Reference-Format}
\bibliography{main}
\end{document}

%% file: abstract.tex
\begin{abstract}
% problem & importance  
Data programming is a programmatic weak supervision approach to efficiently curate large-scale labeled training data. Writing data programs (labeling functions) requires, however, both programming literacy and domain expertise. Many subject matter experts have neither programming proficiency nor time to effectively write data programs. Furthermore, regardless of one's expertise in coding or machine learning, transferring domain expertise into labeling functions by enumerating rules and thresholds is not only time consuming but also inherently difficult. 
% proposed solution 
Here we propose a new framework, data programming by demonstration (DPBD), to generate labeling rules using interactive demonstrations of users. DPBD aims to relieve the burden of writing labeling functions from users, enabling them to focus on higher-level semantics such as identifying relevant signals for labeling tasks. 
We operationalize our framework with \system, an interactive system that synthesizes labeling rules for document classification by using span-level annotations of users on document examples. 
% evidence that it works (solves the problem)
We compare \system with conventional data programming  through a user study conducted with 10 data scientists creating labeling functions for sentiment and spam classification tasks.  We find that \system is easier to use and learn  and offers higher overall satisfaction, while providing discriminative model performances comparable to ones achieved by conventional data programming. 
\end{abstract}

% \begin{abstract}
% Data programming was introduced to reduce the cost of labeled data collection through denoising labeling functions written (manually) by experts with domain knowledge.
% Writing labeling functions manually requires, however, both programming literacy and domain expertise. 
% Transferring domain expertise into labeling functions by enumerating rules and associated thresholds is not only time consuming but also inherently difficult for nontrivial labeling tasks, further inhibiting broader adoption of data programming. 
% In this paper we propose a new framework, data programming by demonstration (DPBD), to interactively learn labeling functions. 
% DPBD aims to relieve the burden of writing labeling functions from users, enabling them to focus on higher-level semantics such as identifying relevant signals for labeling tasks.  We also present two interactive systems, \ruler and \tagruler, that respectively operationalize  the DPBD framework for document classification and tagging tasks.  
% \end{abstract}

%% file: intro.tex
\vspace{-3pt}
\section{Introduction\label{sec:intro}}
Machine learning (ML) models used in practice today are predominantly supervised models and rely on large datasets labeled for training. However, the cost of collecting and maintaining labeled training data remains a bottleneck for training high-capacity supervised models~\cite{sculley2015hidden}. Data programming~\cite{bach2017learning,ratner2017snorkel,ratner2016data} aims to address the difficulty of collecting labeled data by using a programmatic approach to weak supervision by heuristics, where domain experts are expected to provide data programs (labeling functions) incorporating their domain knowledge. Prior work on data programming focuses on modeling and aggregating labeling functions written manually~\cite{ratner2017snorkel,ratner2016data} or generated automatically~\cite{hancock2018training, varma2018snuba} to denoise labeling functions. 
% However, little is known about user experience 
% in writing labeling functions and how to improve it~\cite{wang2019interactive}. 

Writing data programs can be, however, challenging and time consuming.  Most domain experts or lay users have no or little programming literacy, and even for those who are proficient programmers, it is often difficult to convert domain knowledge to a set of rules by writing programs.    

\begin{figure}[tbp]
    \centering
    \includegraphics[width=0.9\linewidth]{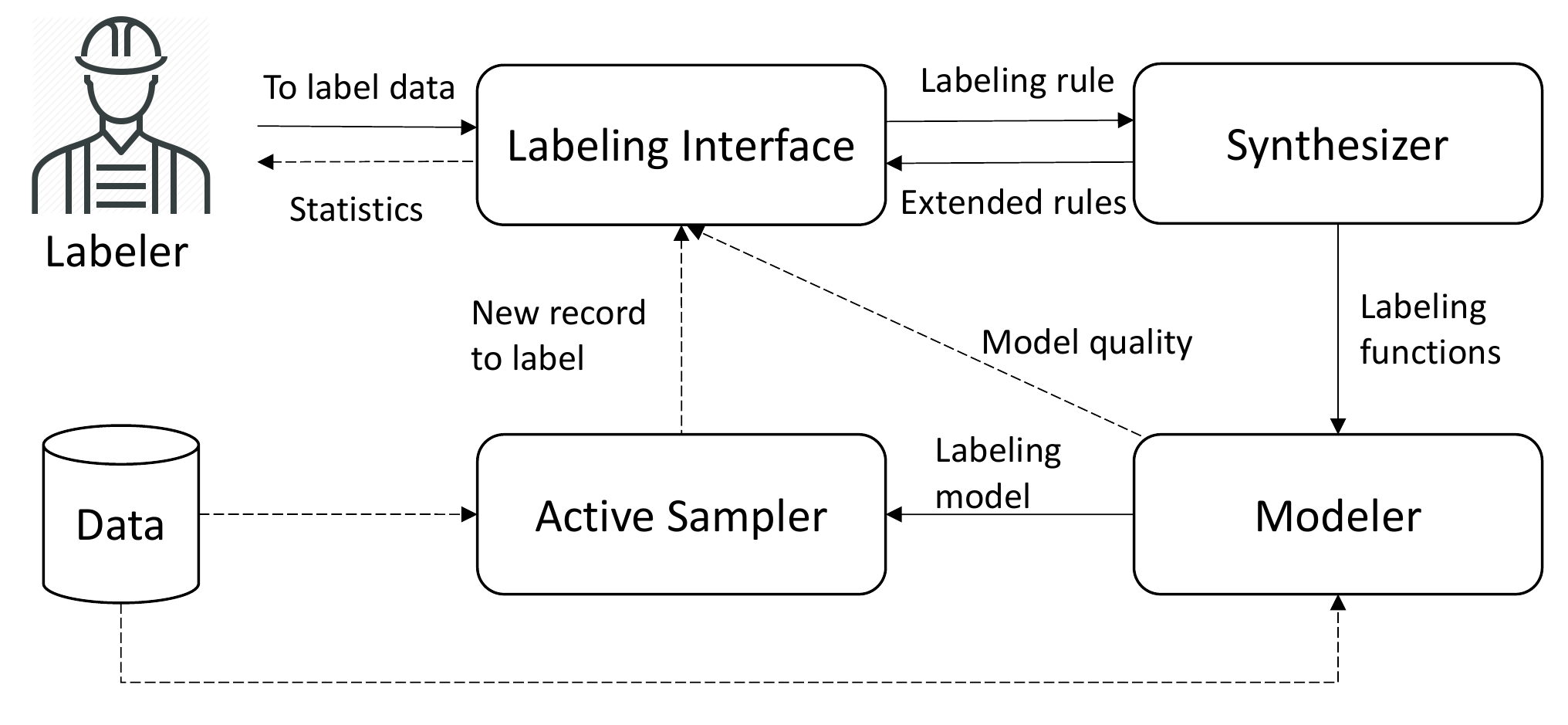}
    \caption{Overview of the data programming by demonstration (DPBD) framework. Straight lines indicate the flow of domain knowledge, and dashed lines indicate the flow of data.\vspace{-10pt}\label{fig:overview}}
\end{figure}
%  By extending data programming with programming by example, we bridge the gap between scalable training data generation and domain experts.
To address these challenges, we introduce data programming by demonstration (DPBD),  a new framework that aims to make creating labeling functions  easier by learning them from users' interactive visual demonstrations. DPBD moves the burden of writing labeling functions to an intelligent synthesizer while enabling users to steer the synthesis process at multiple semantic levels, from providing rationales relevant for their labeling choices to interactively filtering the proposed functions. DPBD draws from two lines of prior research; programming by demonstration (PBD) or example (PBE), e.g.,~\cite{gulwani2011asp,lieberman2001your}, which aims to make programming easier by synthesizing them based on user interactions or input and output examples, and  interactive learning from user-provided features or rationales ~\cite{zaidan2008modeling,zaidan2007using}.  

We operationalize our framework with \system, an interactive system that enables more accessible data programming to create labeled training datasets for document classification. \system automatically generates  document level labeling rules from  span-level annotations and their relations on specific examples provided by users. Through a user study conducted with  10 data scientists, we evaluate  \system alongside manual data programming using Snorkel~\cite{ratner2017snorkel}. We measure the predictive performances of models created by participants for two  common labeling tasks, sentiment classification and spam detection. We also elicit ratings and qualitative feedback from participants on multiple measures, including  ease of use, ease of learning, expressivity, and overall satisfaction.  We find \system facilitates more accessible creation of labeling functions without a loss in the quality of learned labeling  models. 

Tagging or token level classification in text documents is another widely used task that can benefit from DPBD. Here we also briefly discuss our work in progress on \tagruler, a DPBD system that learns token labeling functions through user interaction to create training datasets for tagging models.  
% Tagging or span-level classification in text documents is another widely used task that can benefit from DPBD. Here we also briefly discuss our work in progress on \tagruler, a DPBD system that enables the interactive generation of token labeling functions in order to create labeled training data for tagging models.      
% On the other hand, \tagruler synthesizes token classification (e.g., named entity recognition, opinion extraction, etc.) rules based users. 

In summary, we contribute (1) DPBD, a general data independent framework for learning labeling rules by interactive demonstration; (2) \system, an interactive system operationalizing our framework for document classification tasks; and (3) a comparative user study conducted with data scientists in performing real world tasks to evaluate \system and conventional data programming. We have made our research artifacts, including the \system code and demo, publicly available~\footnote{\url{https://github.com/megagonlabs/ruler}}. 

%  along with the materials and anonymized results of the user study
% \begin{figure}[tbp]
%     \centering
%     \includegraphics[width=0.9\linewidth]{figures/ruler_teaser.png}
%     \caption{\system enables the user to interactively 
%     generate a diverse set of labeling functions through simple, non-programmatic text annotations. Dynamically updated statistics allow the user to quickly test and evaluate ideas.\vspace{-10pt}\label{fig:my_label}}
% \end{figure}

%% file: framework.tex
\section{DPBD Framework}\label{sec:framework}
\stitle{Problem Statement}
Given a dataset $D=\set{d_1, \ldots, d_m}$ of data records and a set of labels $L=\set{l_1,\dots, l_n }$,  we aim to develop a framework that enables human labelers to interactively assign a label from $L$ for each data record efficiently sampled from  $D^\prime \subset D$ ($|D^\prime| \ll |D|$), while demonstrating their rationales for label assignments through visual interaction. Given a triplet $(d_i^\prime,  v_i, l_j)$ of a data record, a visual interaction from the labeler, and the label assigned,  we want this framework to effectively synthesize and propose labeling rules $R_{ij} = \set{r_1,\ldots,r_k}$ for the labeler to choose from. 
Finally, we want the framework to optimally aggregate all the chosen rules (labeling functions) in order to create a labeled training set from  $D\setminus D^\prime$ with probabilistic labels in order to subsequently train discriminative models on it.

\stitle{Framework Overview}
The data programming by demonstration (DPBD) framework  (Figure~\ref{fig:overview}) has two input sources: the human \emph{labeler}, and the \emph{data} that is to be labeled. The labeler is the subject matter expert who has sufficient domain understanding to extract useful signals from data. Given a dataset, our framework enables the labeler to label each record with a categorical label, while providing their labeling rationales by interactively marking relevant parts of the record and specifying semantics and relationships among them.
The output is a labeling model, which is trained to automatically produce labels for the large set of unlabeled data.
The DPBD framework has four main components, \emph{labeling interface}, 
\emph{synthesizer}, \emph{modeler}, and \emph{active sampler}. 
% The labeler interacts with data via the \emph{labeling interface}. The labeling interface 
% records the labeler's interaction and compiles the interaction into a labeling rule. The 
% \emph{synthesizer} synthesizes labeling rules and translates those chosen by the labeler into % program functions. Third, the selected functions are passed to the \emph{modeler}, which 
% builds a labeling model by optimally aggregating the generated functions. Until a certain 
% stopping criterion is met (e.g., reaching a desired model quality) or the labeler decides to exit, % the \emph{active sampler} selects the next data record to present the labeler.  
% In the rest of this section, we describe the details of each component.

\subsection{Labeling Interface}\label{sec:interface} The labeling interface is the workplace where the labeler encodes domain knowledge into labeling rules. It provides a way to express noisy explanations for labeling decisions using a visual interaction language,  which allows the user to express domain knowledge without having to formalize their ideas into computer programs or natural language explanations. This allows for more focus on patterns in the data while abstracting away any implementation concerns.

\stitle{Generalized Labeling Model} Inspired by the entity-relationship model~\cite{DBLP:journals/tods/Chen76} in database modeling, the generalized labeling model (GLM) models the data records with \emph{concepts} and \emph{relationships}.  The GLM views the data record as a series of tokens,
where a token is a continuous subset of a record with no semantics attached. 
For example, in text data, a token can be any span (single char to multiple words) of the data record; in an image data record, it would be a 2D region, rectangular or free form; and in an audio data record, it would be a 1D window of the data record (e.g., a phoneme).  A \emph{concept} is a group of tokens that the labeler believes share common semantics. For instance, over text data, the labeler might define a concept of positive adjectives consisting of a set of tokens, each of which can imply a positive review. 
When labeling audio data, the labeler might create a concept to aggregate all clips that express excitement, or of a specific speaker.  This abstraction allows the user to teach the GLM which generalizations are relevant to the task.
A \emph{relationship} represents a binary correlation between token-token, token-concept, or concept-concept. Some examples are membership (e.g., a token is in a concept), co-existence (e.g., opinion and aspect tokens), and positional (e.g., a person is standing left to a table~\cite{DBLP:conf/ijcnn/HaldekarGO17}).

\begin{table}[h]
\centering
  \caption{Mapping from GLM elements to operations in the labeling interface.\label{tab:GLM}}
  \begin{tabular}{cc}
    \toprule 
    GLM Element & Operations\\
    \midrule
     token & select, assign\_concept \\
     concept & create, add, delete\\
     relationship & link, direct\_to\\
  \bottomrule
\end{tabular}
\end{table}

\stitle{Mapping GLM Elements to Operations}
Given the GLM specification described above, our framework also defines the operations that can be applied on GLM elements. Table~\ref{tab:GLM} lists the GLM elements and the corresponding operations. The implementation of both the labeling interface and the operations described in Table~\ref{tab:GLM} would vary across data types and token definitions. To add expressivity, the GLM may also perform transformations over the set of tokens, as we describe in the next section. 

\stitle{Compiling Operations into Labeling Rules}
Once the labeler finishes annotating an example using the provided operations, and selects a label, the tokens are extracted from the annotation and used 
as the initial set of conditions from which to build rules.
The synthesizer combines these conditions into labeling rules by selecting subsets of the conditions to be combined with different conjunctive formulas, according to the relationships the user has annotated.
The synthesizer extends the initial set of labeling rules and presents the extended labeling rules for the labeler to select from, choosing desired ones based on  domain knowledge.

A labeling rule serves as an intermediate language, interpretable by both the labeler and the synthesizer. In our framework, we adopt the notation of domain relational calculus~\cite{DBLP:conf/vldb/LacroixP77} to represent these rules, which can be expressed as: 
$\{\texttt{tokens} \mid \texttt{conditions}\}\Rightarrow  \texttt{label} $.
The variable \texttt{tokens} is a sequence of tokens with existential quantification, and
 \texttt{conditions} is a conjunctive formula over boolean predicates that is tested over \texttt{tokens} on a data record.
 
The predicates are first-order expressions, and each can be expressed as a tuple $(T, lhs, op, rhs)$. $T$ is an optional transformation function on a token identifier, a process of mapping the raw token to more generalized forms. Some example transformations are word lemmatization in text labeling, speech-to-text detection in audio labeling, or object recognition in image labeling.  
$lhs$ is a token, while $rhs$ is can be either token, literal or a set.
If $rhs$ denotes a token, the transformation function $T$ may also apply to $rhs$. 
$op$ is an operator whose type depends on the type of $rhs$.  If $rhs$ is a token or literal,  $op$ detects a positional or an (in)equality relationship. Otherwise, if $rhs$ is a set, $op$ is one of the set operators  $\{\in, \not\in \}$. Since the \texttt{conditions} is in the conjunctive form, the order of labeler's interactions does not matter.

\example{
Consider the following review for the binary sentiment classification (positive or negative) task:

\texttt{This book was so great! I loved and read it so many times that I will soon have to buy a new copy.} 

If the labeler thinks this data record has a positive sentiment, she can express her decision rationale using GLM.
First, she may select two tokens that are related to the sentiment: \texttt{book} and \texttt{great}. Assume there are two concepts the labeler previously created: (1) \texttt{item$=\{$book, electronics$\}$}; and
(2) \texttt{padj$=\{$wonderful$\}$}. The labeler realizes the token \texttt{great} can be generalized by the \texttt{padj} concept, which means that the labeling rule will still be valid if this token is replaced by any tokens in the concept, so she adds this token to the concept.

Finally, the labeler creates a positional relationship from \texttt{book} to token \texttt{great} to indicate that they appear in the same sentence, before completing the labeling process.
These operations compile into the labeling rule
$r: \{t_1, t_2 \mid t_1=\texttt{book} \land t_2 \in \texttt{padj} \land idx(t_1) < idx(t_2)  \}\Rightarrow \texttt{positive}$.
}

This rule is sent to the synthesizer for expansion and program synthesis.

\subsection{Synthesizer}\label{sec:synthesizer}
Given the compiled labeling rule from the labeling interface, the synthesizer extends one single labeling rule from labeler's interaction to a set of more general labeling rules; and translates those labeling rules into computer programs.
It is straightforward to translate the rules into executable computer programs (labeling functions), so in this section, we focus on how to synthesize the extended labeling rules.

Given the labeling rule compiled from a labeler's interaction, the synthesizer generates more labeling rules while optimizing two competing goals: maximizing generalization, so that more (unseen) data can be accurately labeled; and maximizing the coverage of the labeler's interaction, simply because labeler's interaction is the most valuable signal for labeling based on the domain knowledge.
Of course, the larger the set of annotations in an interaction, the larger the set of labeling functions that can be synthesized. To keep rule selection as easy as possible for the user, in this case we prioritize rules that cover more of the interaction, assuming that there is little redundancy.% in the labeler's interaction.

We achieve generalization of the given rules using the following heuristics:
(1) substituting tokens with concepts; (2) replacing general coexistence relationships with position-specific ones; and (3) applying the available transformations over the tokens (for example, object recognition in a section of an image).

% Since the labeling rule in GLM has conjunctive conditions, Algorithm~\ref{alg:synthesizer} (Line~\ref{alg:syn_begin}-\ref{alg:syn_end}) generalizes each predicate in the conditions. 
% Inside, Line~\ref{alg:syn_findset} to Line~\ref{alg:syn_h1_end} substitute token with concept.
% Line~\ref{alg:syn_findset} can be implemented explicitly by matching token to concept set, as well as sophisticated data-dependent processing via transformation .
% For example, in our system for text labeling (Section~\ref{sec:ruler}), in addition to matching values with labeler defined concepts, we also apply named-entity recognition (NER) where the named-entities are implicit concepts that a token can be a member of. 
% Line~\ref{alg:syn_h2_begin} to Line~\ref{alg:syn_h2_end} replace the positional with co-occurrence relationship by removing the condition that specifies the positional context.
% The conditions for extended labeling rules is a conjunctive combination of single predicates, one from each extended condition set (Line~\ref{alg:syn:combo}).
% In addition, for special cases of binary labeling, the algorithm also considers the rule which flips over the label by adding negation to the conditions (Line~\ref{alg:syn_h3}).

%\input{lf_extension_algorithm}

Once the extended rules are generated, the rules are ranked by their generalization score---a measurement of how applicable a certain rule is.
We define a data-independent generalization score for a labeling rule $r$ as: $ G(r) = \prod_{c\in r.conds} |c.rhs|$. 
% \begin{equation*}
% G(r) = \prod_{c\in r.conditions} |c.rhs|
% \end{equation*}
Intuitively, $G(r)$ is calculated by counting how many different data instances that $r$ can be used.
%It prefers labeling rules using large sets to match tokens in the data record.

\example{
Continuing with our Amazon review example, the synthesizer can derive the following labeling rules from $r$ using these heuristics:
\begin{enumerate}
\small{
\item $\{t_1, t_2 \mid t_1 \in \texttt{item} \land t_2 \in \texttt{padj} \} \Rightarrow \texttt{positive}$\label{rule1}
%\item $\{t_1, t_2,t_3 \mid t_1 \in \texttt{item} \land t_2 \in %\texttt{padj} \land t_3 \in %\texttt{neg}\}\Rightarrow\texttt{negative}$\label{rule2}
\item $\{t_1, t_2 \mid t_1 \in \texttt{item} \land t_2 \in \texttt{padj} \land idx(t_1)<idx(t_2)\}\Rightarrow \texttt{positive}$\label{rule3}
%\item $\{t_1, t_2, t_3 \mid t_1 \in \texttt{item} \land t_2 \in %\texttt{padj} \land idx(t_1)<idx(t_2) \land t_3 \in %\texttt{neg}\}\Rightarrow\texttt{negative}$\label{rule4}
\item $\{t_1, t_2 \mid t_1=\texttt{book} \land t_2\in \texttt{padj} \}\Rightarrow \texttt{positive}$\label{rule5}
%\item $\{t_1, t_2, t_3 \mid t_1=\texttt{book} \land t_2\in %\texttt{padj} \land t_3 \in \texttt{neg} \}\Rightarrow %\texttt{negative}$\label{rule6}
}
\end{enumerate}
%Labeling rule~(\ref{rule1}) is generated using heuristics (1) and (2).  Labeling rule~(\ref{rule3}) and~(\ref{rule5}) are synthesized by using heuristics (1) and (2), respectively.
Note that labeling rule~(\ref{rule1}) is more general than~(\ref{rule3}) and~(\ref{rule5}) because all data records that can be labeled by~(\ref{rule3}) and~(\ref{rule5}) will be labeled the same way using labeling rule~(\ref{rule1}).
%Labeling rules~(\ref{rule2},~\ref{rule4},~\ref{rule6}) are due to flipping over the binary label with heuristics (3). 
}

%Once the extended labeling rules are generated, the labeler can help confirm the validity in order to achieve faster convergence.
The top-k candidates ranked by the generalization score are displayed in the labeling interface for the labeler to accept or reject.

\subsection{Modeler}\label{sec:modeler}
The modeler component trains a model that can be used to automatically annotate unlabeled datasets.
Naively aggregating the labeling functions  can be either inaccurate (since labeling functions can be conflicting and correlated), or does not scale with a large set of unlabeled data~\cite{ratner2017snorkel}. Instead, the modeler encapsulates the ideas from traditional data programming~\cite{ratner2019training, bach2017learning, ratner2017snorkel,ratner2016data} to first build a generative model to denoise labeling functions, and then train a discriminative model to leverage other features beyond what are expressed by the labeling functions.

\subsection{Active Sampler}\label{sec:sampler}
To improve the model quality at faster rates, our framework uses an active sampler to choose the next data record for labeling.  The active sampler can be plugged in with any custom active learning policy.  By default, it selects the data record $x^*$ with the highest entropy (i.e., the one that the labeling model is currently  the most uncertain about): $ x^* = \argmax_x - \sum_i^{|L|} p_\theta(L_i\mid x)\log{p_\theta(L_i\mid x)}$  where $p_\theta(L_i\mid x)$ is the probability that example $x$ belongs to class $L_i$, as predicted by the trained label model.

%% file: ruler.tex
\begin{figure}[!t]
    \centering
    \includegraphics[width=\linewidth]{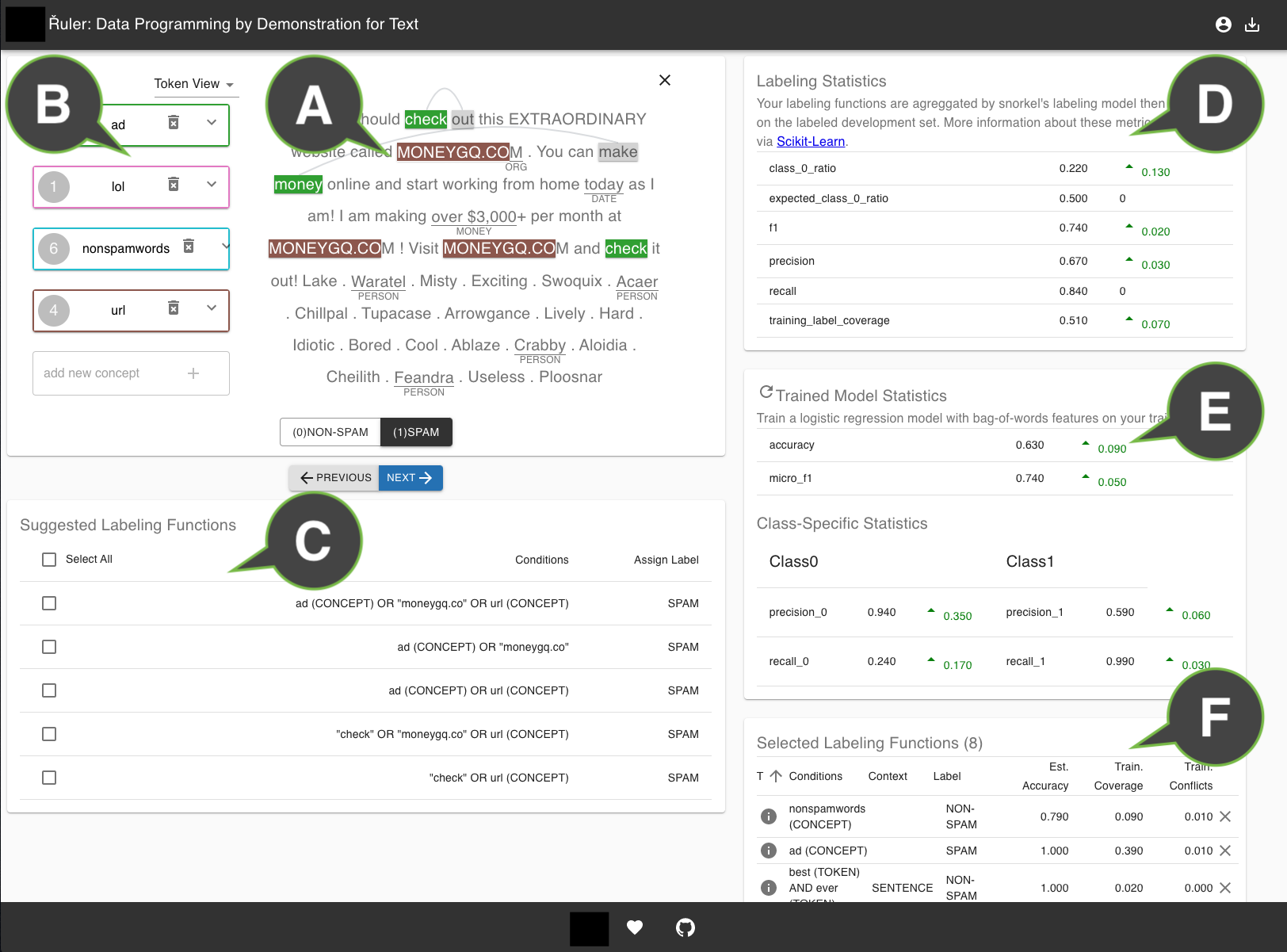}
    \caption{\system user interface. \system synthesizes labeling  rules based on rationales expressed by users by interactively marking relevant parts of the example and specifying implied  group or pairwise semantic relations among them.  \label{fig:ruler}}
 \end{figure}

\section{Ruler}\label{sec:ruler}
\system is an  interactive system that implements the data programming by demonstration (DPBD) framework to  facilitate labeled training data preparation for document-level text classification models. For this, \system leverages   span-level features and relations in text documents demonstrated  through visual interactions by users (labelers). To begin a labeling task, the data owner needs to upload their unlabelled dataset, in addition to a small labeled development set, and optionally a small test and validation set.  This mirrors the data requirements of Snorkel,  which the underlying DPBD modeler encapsulates. We next discuss the user interface and interactions of \system along with its implementation details in operationalizing DPBD for text.  

\subsection{User Interface and Interactions}
Recall that the purpose of the labeling interface in DPBD ( Section~\ref{sec:interface}) is to enable the labeler to encode domain knowledge into rules through visual interaction. To this end, \system interface provides affordances through 6 basic views (Figure~\ref{fig:ruler}), which we briefly describe below---the letters A-F refer to annotations in Figure~\ref{fig:ruler}. 

\begin{figure}[tbp]
    \centering
    \includegraphics[width=\linewidth]{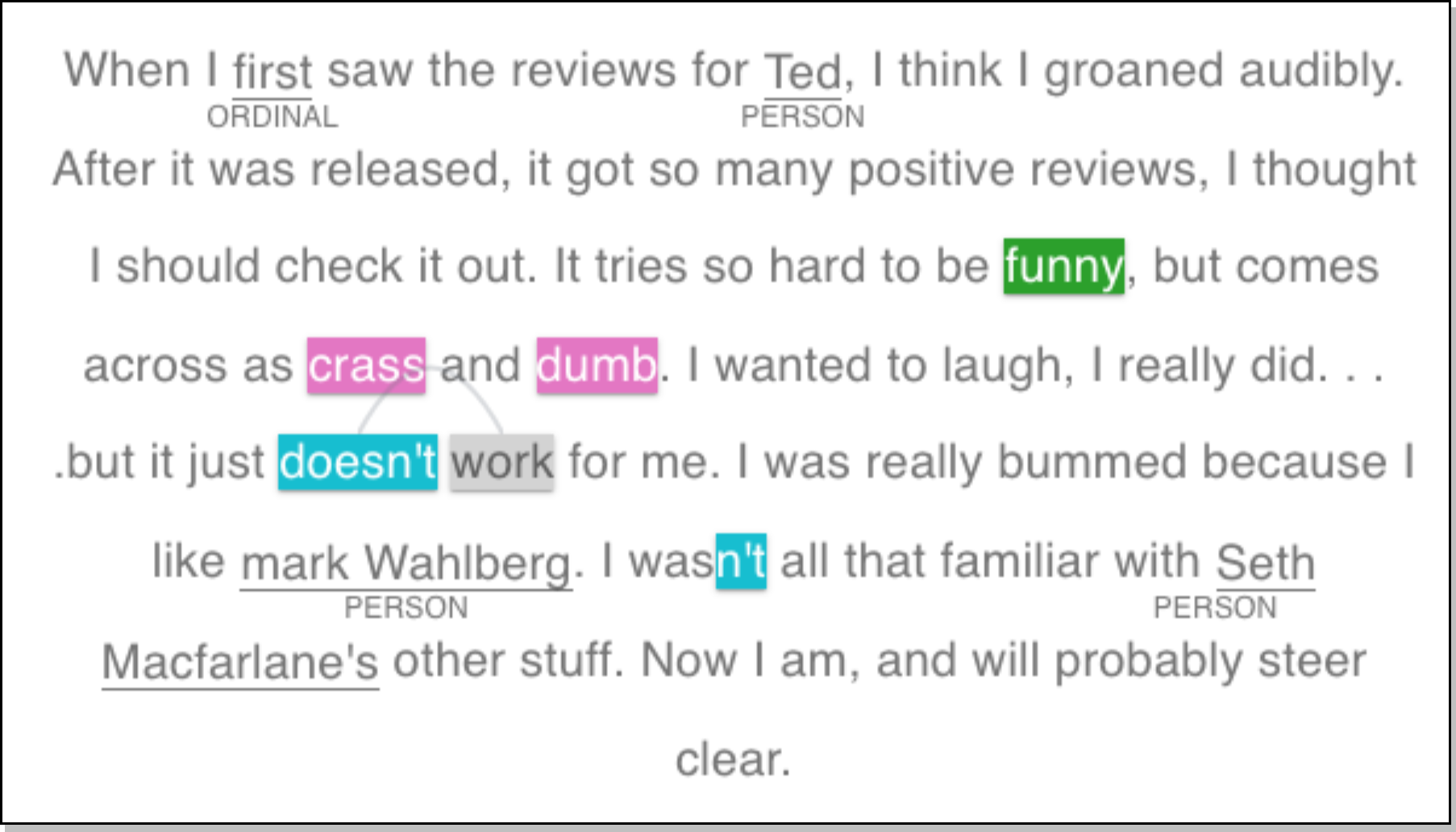}
    % \vspace{0pt}
    \caption{\system Labeling Pane, where the user conveys domain knowledge using a visual interaction language. Annotations are color coded by the concepts they are assigned to.\vspace{-5pt}\label{fig:labeling_pane}}
\end{figure}

\subhead{Labeling Pane} (A) is the main view where the user interacts with document text. Labeling Pane (Figure~\ref{fig:labeling_pane}) shows  contents of a single document at a time and supports all the labeling operations defined by the GLM in the context of text data.  The user can annotate spans either by highlighting them directly with the cursor or adding them to a concept. These spans can be linked together if the relationship between them is significant to the user. Once the user selects a document label (class)  from the options displayed, the system synthesizes a diverse set of labeling functions to suggest to the user. 
    
\begin{figure}[tbp]
    \centering
    \includegraphics[width=\linewidth]{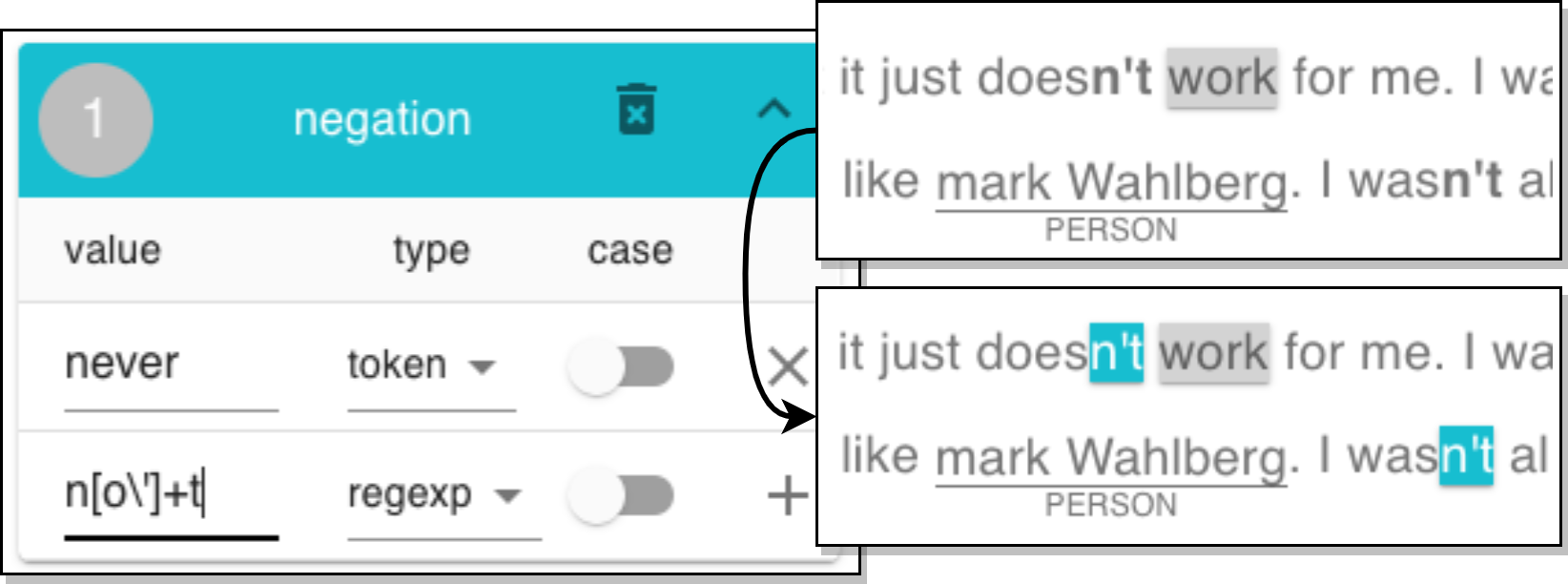}
    % \vspace{0pt}
    \caption{Left: Example concept created to capture negation.  Right: example text highlighting as concept elements are matched in the text, and annotations created once the element is submitted.\label{fig:concepts_pane}}
\end{figure}

\subhead{Concepts Pane} (B) allows users to create concepts, add and edit tokens (whole words surrounded by non-alphabetical characters) or regular expressions, and see annotations over their text automatically added when a match is found (Figure~\ref{fig:concepts_pane}). This interaction allows users to abstract away details about specific language use by grouping tokens or regular expressions into  concepts.

\subhead{Suggested Functions} (C) shows the labeling functions suggested by the system. The user can select any functions that seem reasonable, and only then are they added to the underlying labeling model that is iteratively built. 

\subhead{Labeling Statistics} (D) displays current statistics of the label model computed over the development set, and differential changes incurred by the last data interaction. Because this panel updates as the user interacts, they can quickly explore the space of labeling functions with a very low cost in terms of time, computation, and human effort.

\subhead{End-model Statistics} (E) shows the performance statistics for an end model for which the user intends to collect training data. For example, in our user study we used a logistic regression model  with a bag of words features on the generated training data. This model is evaluated on a small held-out test set, and its performance metrics are shown in this pane.

\subhead{Selected Functions} (F) lists of currently selected labeling rules that make up the labeling model along and shows each rule's performance statistics based on the development set. The user can click to open a details panel showing observed incorrect labels and sample texts labeled by this function.

\subsection{Server and Model}
\system's backend comprises the synthesizer (Section~\ref{sec:synthesizer}), modeler (Section~\ref{sec:modeler}) and active sampler (Section~\ref{sec:sampler}) components. The backend components are all implemented in Python 3.6. 

In addition to the function generation defined in Section~\ref{sec:synthesizer}, \system's synthesizer also augments labeling rules using existing text processing libraries. It enhances the text with transformations that recognize named entity types such as \texttt{person} and \texttt{location}, extracted using the spaCy library~\cite{spacy}. These annotations are made visible to the user, and annotations containing named entities will generate functions that generalize to all instances of that entity. 

In our implementation, relationships can include co-occurrence in the same sentence as well as in the document.
\system encapsulates the Snorkel library~\cite{ratner2017snorkel} into its modeler to aggregate the generated labeling functions. 

%% file: evaluation.tex
\section{Evaluation}\label{sec:exp}
We evaluate \system alongside manual data programming using Snorkel~\cite{ratner2017snorkel}. Our goal is to better understand the trade-offs afforded by each method. To this end, we conducted a user study with data scientists and measured  their task performance accuracy in completing two labeling tasks. In addition to task performance, we also analyzed the accessibility and  expressivity of both methods using the qualitative feedback elicited from participants. 

Note that \system can be used by programmers and non-programmer domain experts alike, but a fair comparison with Snorkel requires proficiency in conventional programming. 

\subhead{Participants}  We recruited 10 participants with Python programming experience through our professional network. All participants had significant programming experience (avg=$12.1$ years, std=$6.5$). Their experience with Python programming ranged from $2$ to $10$ years with an average of $5.2$ years (std=$2.8$).  

\subhead{Experimental Design}  We carried out the study using  a within-subjects experiment design, where all participants performed tasks using both conditions (tools).  The sole independent variable controlled was the method of creating labeling functions. We counterbalanced the order in which the tools were used, as well as which classification task we performed with which tool. 

\subhead{Tasks and Procedure} We asked participants to write  labeling functions for two prevalent labeling tasks: spam detection and sentiment classification.  They performed these two tasks on  YouTube Comments and Amazon Reviews, respectively. Participants received 15 mins of instruction on how to use each tool, using a topic classification task (electronics vs. guns) over a newsgroup dataset~\cite{rennie200820} as an example. We asked participants to write as many functions as they considered necessary for the goal of the task.  There were given 30 mins to complete each task and we recorded the labeling functions they created and these functions' individual and aggregate performances.  After completing both tasks, participants also filled out an exit survey, providing their qualitative feedback.

For the manual programming condition, we provided a Jupyter notebook interface based on the Snorkel tutorial. The notebook had a section for writing functions, a section with diverse analysis tools, and a section to train a logistic regression model on the labels generated. 

%% file: results.tex
\begin{figure}[tbp]
\centering
\includegraphics[width=\linewidth]{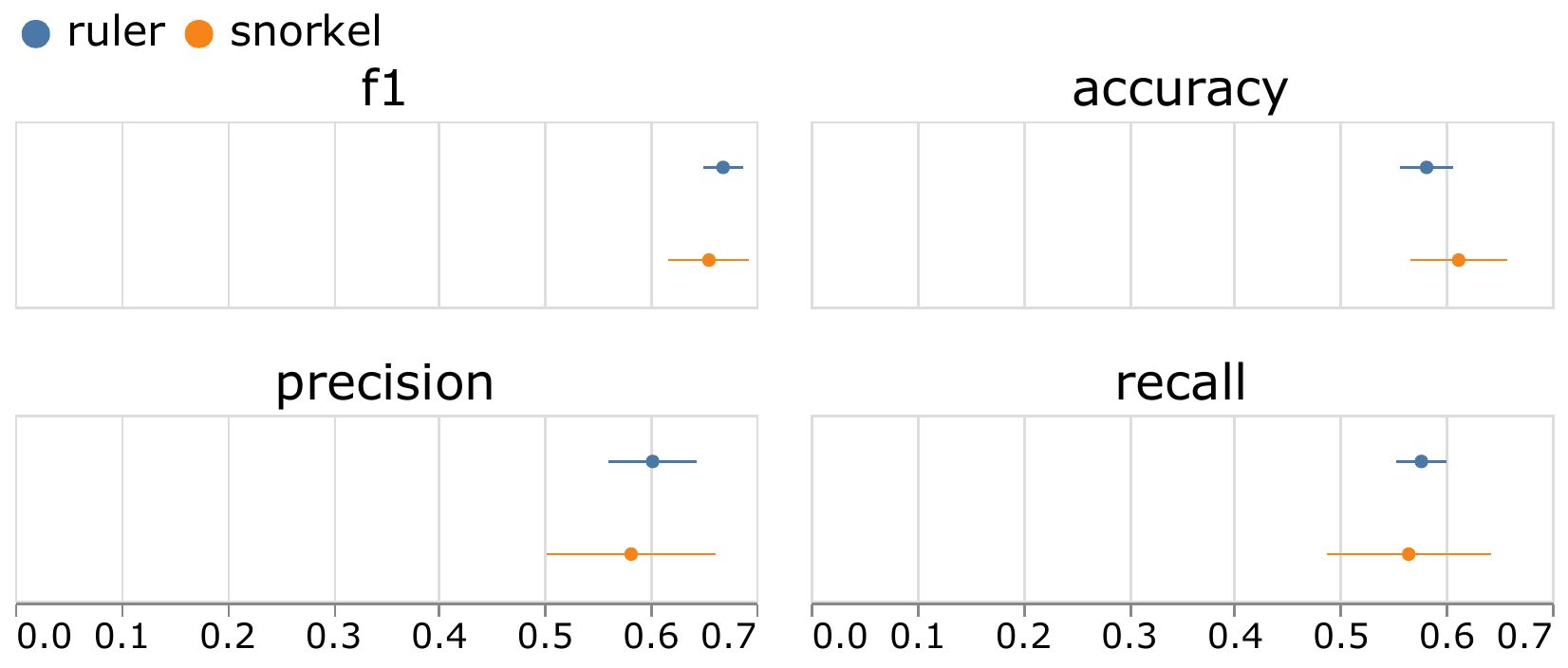}
\caption{Performances of the classifier models trained on the probabilistic labels generated by participants' labeling models. Although manual programming allows participants to use existing packages (e.g., sentiment analysis packages), \system performs comparably with Snorkel in both tasks.
\vspace{-10pt}\label{fig:modelperfm}}
\end{figure}
\section{Results}\label{sec:results}
To analyze the performance of the labeling functions created 
by participants, for each participant and task, we select the labeling  
model  that achieved the highest f1 score on the development set.  For each labeling model, we then train a logistic regression model on a training dataset  generated by the model.  We finally evaluate the performance of the logistic regression model on a heldout test set (400 examples). We also analyze the subjective ratings provided by participants on a Likert scale of 5 (1--5, higher is better)  in their exit surveys.  
We use the paired Wilcoxon signed rank test to assess the significance of differences in prediction metrics and subjective ratings between \system and Snorkel. We also report the effect size $r$ for all our statistical comparisons.   

%Figure \ref{fig:time_elapsed} shows the distribution of the times until the best model was created. 
%Although users created many more labeling functions using \system (see \ref{table:mean_values},) the spread of elapsed times until the best model is achieved is similar to that for Snorkel, suggesting that the increased number of functions does not cause users to plateau sooner.

\subhead{Model Performance}
We find that \system and Snorkel provide comparable model performances (Figure~\ref{fig:modelperfm}). The logistic regression models trained on data produced by labeling models created using \system have slightly higher f1 ($W=35$, $p=0.49$, $r=0.24$ ), precision ($W=30$, $p=0.85$, $r=0.08$), and recall ($W=25$, $p=0.85$, $r=0.08$) scores on average. Conversely, accuracy is slightly higher ($W=17$, $p=0.32$, $r=0.15$) for Snorkel models on average than \system. However these differences are not statistically significant. 

\subhead{Subjective Ratings and Preferences} 
Participants find \system to be significantly easier to use 
($W=34$, $p=0.03 < 0.05$, $r=0.72$) than Snorkel (Figure~\ref{fig:subjectiveratings}). 
Similarly, they consider \system easier to learn ($W=30$, $p=0.1$, $r=0.59$) than Snorkel.  On the other hand, as we expected, participants report Snorkel to be more expressive ($W=0$, $p=0.05$, $r=0.70$)  than  \system. However, our participants appear to consider accessibility (ease of use and ease of learning) to be more important criteria, rating \system higher ($W=43$, $p=0.12$, $r=0.51$) than Snorkel for overall satisfaction.  
%  When asked which tool they prefer overall, 2 users prefered Snorkel, 4 prefered \system, and the remaining 4 said it depends on the task and data. If they wanted to get data quickly, or if the dataset required a lot of domain-specific keywords, many would opt for \system, whereas Snorkel would be useful given more time.  
% One user summarized it thus ``Simple label function[s] that rely on keywords are much easier and faster to write with \system. For both tasks, I did not write complex label logic, so with the same time, I can write more label functions with \system.'' 
% The reason users preferred Snorkel in certain situations was for added expressivity, yet interestingly almost three-quarters ($72.3\%$) of the functions that users wrote in Snorkel could be captured through \system interactions.  The types of functions not captured included: those using Python sentiment analysis packages, and functions that counted the number of occurrences of a word, the length of the text, or, in one case, the ratio of alphabetical characters.  This suggests that even skilled programmers can benefit from using both systems, using \system to more quickly capture domain specific concepts and language use, and then manually adding functions based on their new understanding of the data.  
These results show that \system is more accessible and offers better overall experience, while providing model performance comparable to Snorkel. \system can help data scientists save time and create better models, either in conjunction with traditional programming or alone. For the user who is not skilled at programming, \system is also the only tool available to help leverage data programming with control over the functions.

%% file: related.tex
\vspace{-5pt}
\section{Related Work}\label{sec:related}
We build on prior work on weak supervision, programming by demonstration, and learning from feature annotations by users.

\subhead{Weak Supervision}\label{sec:weak_supervision}
To reduce the cost of labeled data collection, weak supervision methods leverage noisy, limited, or low precision sources such as crowdsourcing~\cite{karger2011iterative}, distant supervision~\cite{mintz2009distant}, and user-defined heuristics~\cite{gupta2014improved} to gather large training data for supervised learning. Data programming~\cite{ratner2017snorkel,ratner2016data} is a programmatic approach to weak supervision using heuristics, where labeling functions provided by domain experts are used to create training data at scale.  
%  and train ML models using probabilistic labels.

\subhead{Program Synthesis by Demonstration}\label{sec:program_synthesis}
Automated synthesis of programs that satisfy a given specification is a classical artificial intelligence (AI) problem~\cite{waldinger1969prow}. Generating programs by examples or demonstration is an instance of this problem. The terms programming by example (PBE), or programming by demonstration (PBD) have often been used interchangeably, though their adoption and exact meaning might diverge across fields and applications.  PBD systems aim to empower  end user programming in order to improve user productivity
~\cite{allen2007plow,leshed2008coscripter,li2017sugilite,myers1998topaz,hanafi2017seer,heer2015predictive,jin2019clx}. 
% ~\cite{ allen2007plow,dibia2019data2vis,leshed2008coscripter,li2017sugilite,little2007koala,myers1993peridot,myers1998topaz}. 
One of the core research questions in PBD is how to generalize from seen examples or demonstrations. To generalize, PBD systems need to resolve the semantic meaning of user actions over relevant (e.g., data) items.  Prior approaches incorporate a spectrum of user involvement, from making no inference (e.g.,~\cite{halbert1993smallstar, myers1998topaz}) to using AI models with no or minimal user involvement, to synthesize a generalized program (e.g.,~\cite{gulwani2011asp,lau2003programming,mcdaniel1999getting,menon2013machine,jin2019clx}). Our framework takes a hybrid approach within the spectrum above and combines inference and statistical ranking along with interactive demonstration. 

\subhead{Learning from Feature Annotations}\label{sec:feature_annotation}
Prior  work proposes methods for learning from user provided features~\cite{druck2009active,liang2009learning,raghavan2005interactive}, rationales~\cite{arora2009interactive,von2006peekaboom,zaidan2008modeling,zaidan2007using}, and natural language explanations~\cite{hancock2018training,srivastava2017joint}. BabbleLabble~\cite{hancock2018training} uses a rule-based parser  to turn natural language explanations into labeling functions and aggregates these functions using Snorkel. 
\system also learns labeling functions from high level imprecise explanations and aggregates them using the Snorkel framework. However, \system  enables users to supply their rationales through interactive visual demonstrations, reducing the cognitive load to formalize one's intuition.

\begin{figure}[tbp]
\centering
\includegraphics[width=\linewidth]{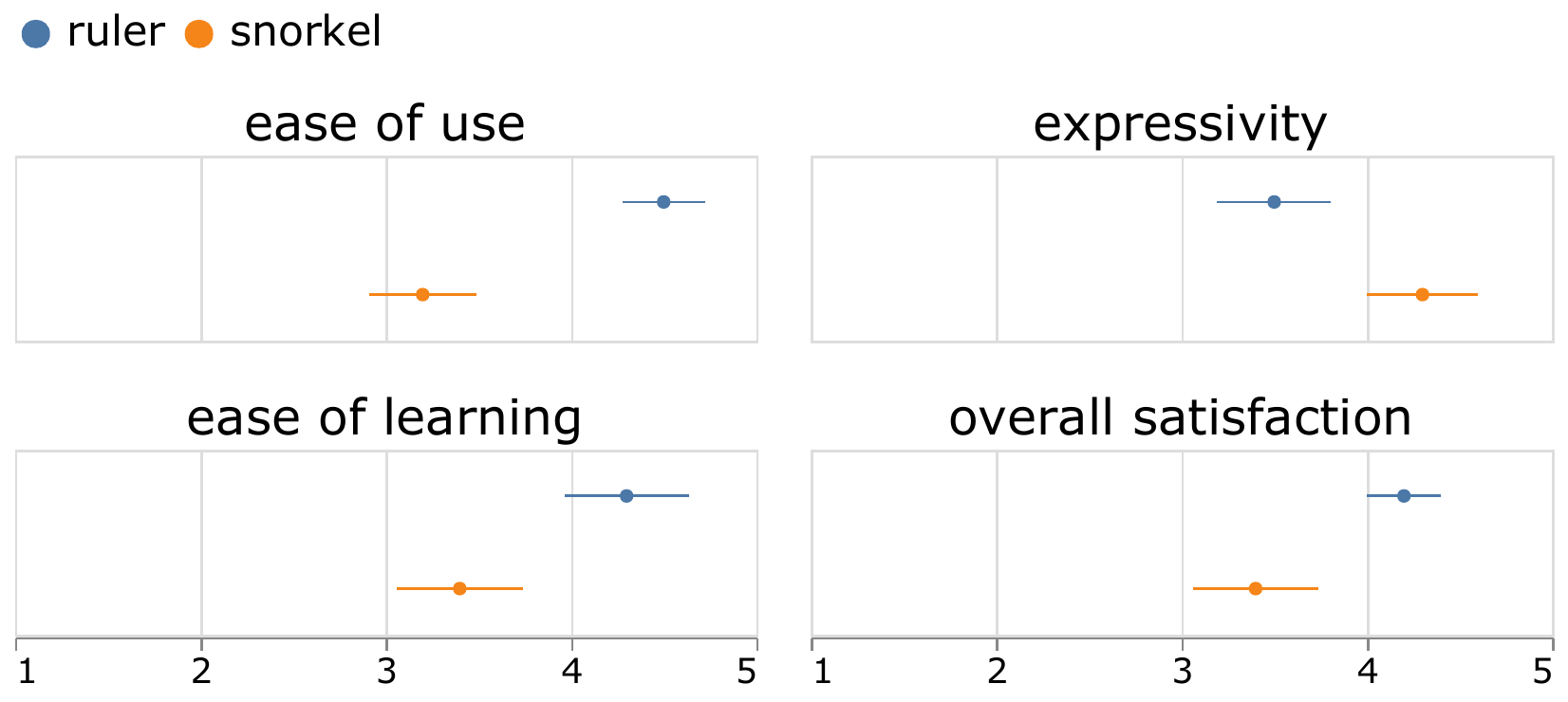} \caption{Participants' subjective ratings on ease of use, expressivity, ease of learning and overall satisfaction, on a 5-point Likert scale. \system is considered by participants to easier to use and learn, though using Snorkel alone is considered to be more expressive.\vspace{-10pt} \label{fig:subjectiveratings}}
\end{figure} 

%% file: tagruler.tex
\section{TagRuler}\label{sec:tagruler}
\begin{figure}[tbp]
    \centering
    \includegraphics[width=\linewidth]{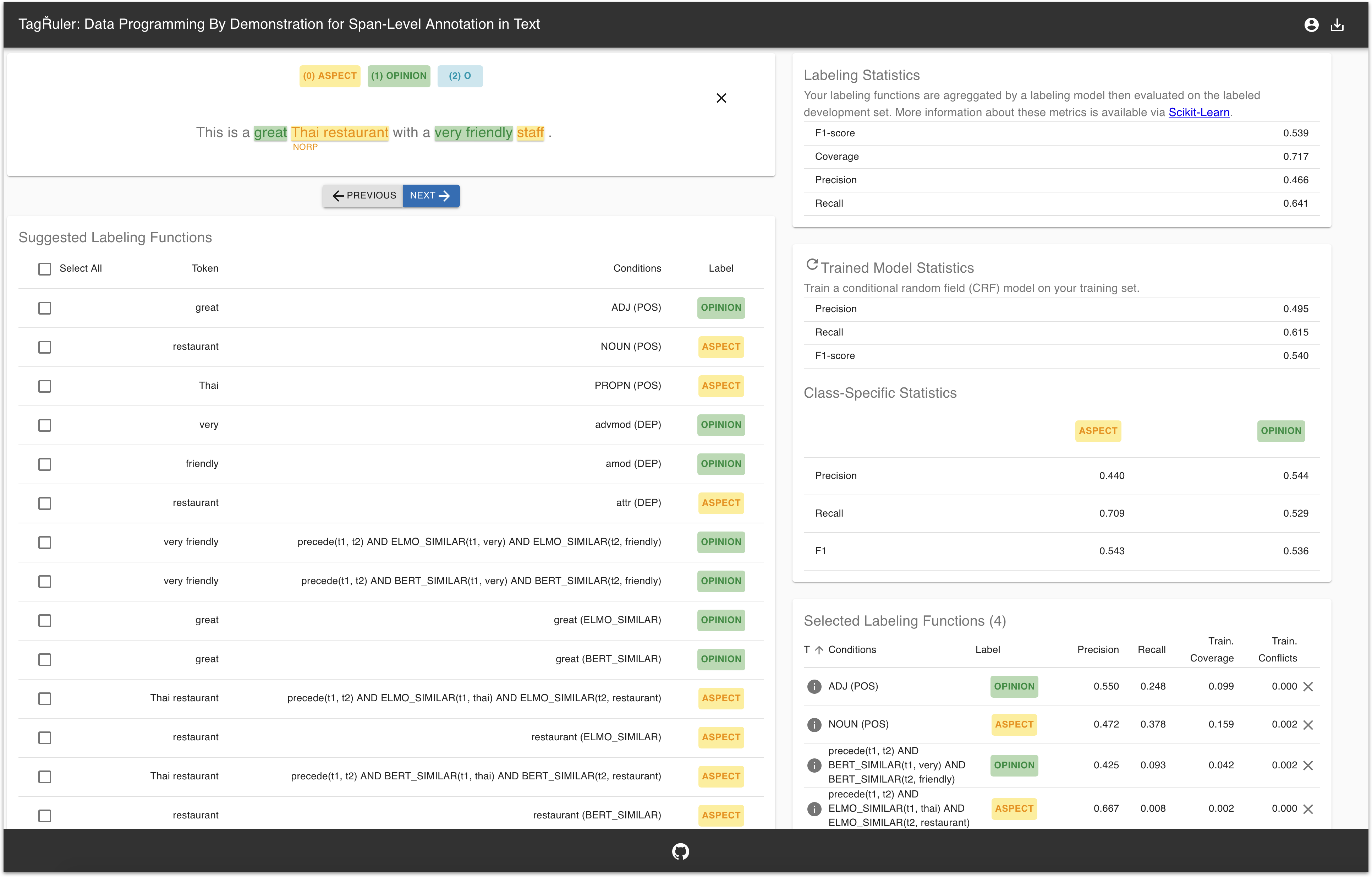}
    \caption{ \tagruler is a DPBD system that synthesizes token level labeling functions based on user annotations.
    \label{fig:tagruler} \vspace{-5pt}}
\end{figure}

Another important task relevant to text documents is tagging, 
the process of classifying token sequences in documents, e.g., as name, person, organization, aspect, opinion, number, etc. Tagging has many applications, including named entity recognition (NER), information extraction, relation extraction, semantic role labeling, and question answering.  Curating training data for tagging models can be a challenge, particularly in domains where experts are needed for data labeling. Based on the DPBD framework, we have been also developing \tagruler (Figure~\ref{fig:tagruler}), an interactive system for learning labeling functions by demonstration for document tagging. While \tagruler shares the underlying framework and user interface characteristics with \system, it tackles a new set of challenges. First, the possibilities for interactive demonstration, where the labeling task itself involves span highlighting, are limited. This makes both intuitive visual interaction design and effective automated inference based on the context of user demonstrations critical. Second, the space of labeling functions can be very large and therefore having  fast and effective synthesis and ranking is crucial. Third, due to the increased number of labeling functions considered, computational performance to sustain interactivity can be a bottleneck.
%Keeping the unit interaction cycle going from user demonstration to synthesized rules and %updating the labeling model and displaying corresponding statistics under  500 ms is a % challenge. 
% The effectiveness of active learning becomes even more important. 

\tagruler currently synthesizes two classes of rules based on the semantic and syntactic analysis of the document context. For semantic analysis, we use the pretrained language models BERT~\cite{devlin2019bert} and ELMo~\cite{peters2018elmo}. As for syntactic analysis, we utilize part of speech (POS) and NER tags along with dependency parsing. To aggregate functions synthesized, the system  provides four alternative models; a hidden Markov model~\cite{safranchik2020weakly}, FlyingSquid~\cite{fu2020fast},  
Snorkel MeTaL~\cite{ratner2019training}, and a majority voting based model. We are currently evaluating the tradeoffs among these alternative  models in the interactive setting of \tagruler.
%  Prior work applies data programming to tackle the challenge~\cite{fries2017swellshark,ratner2019training,safranchik2020weakly}. 

%% file: discussion.tex
\section{Discussion}\label{sec:discussion}
Accessibility is a key to wider adoption of any technology and machine learning is no exception. Here  we introduced  data programming by demonstration (DPBD), a general human-in-the-loop framework that aims to ease writing labeling functions, improving the accessibility and efficiency of data programming.  We then presented \system, a DPBD system, for easily generating labeling functions to create training datasets for document-level classification tasks. \system converts user rationales interactively expressed as span-level annotations and relations among them to labeling rules using the DPBD framework. We also reported our progress in developing \tagruler, a second DPBD system focusing on labeling functions for tagging. 

Through a user study with 10 data scientists performing real world labeling tasks for classification, we evaluated \system together with conventional data programming and found that \system enables more accessible data programming without loss in the performance of labeling models created. We believe DPBD systems will be useful for data scientists as well as subject matter experts. We release \system as open source software to support future applications and extended research.

%% file: main.bbl

\begin{thebibliography}{41}

%%% the Web address.

\ifx \showCODEN    \undefined \def \showCODEN     #1{\unskip}     \fi
\ifx \showDOI      \undefined \def \showDOI       #1{#1}\fi
\ifx \showISBNx    \undefined \def \showISBNx     #1{\unskip}     \fi
\ifx \showISBNxiii \undefined \def \showISBNxiii  #1{\unskip}     \fi
\ifx \showISSN     \undefined \def \showISSN      #1{\unskip}     \fi
\ifx \showLCCN     \undefined \def \showLCCN      #1{\unskip}     \fi
\ifx \shownote     \undefined \def \shownote      #1{#1}          \fi
\ifx \showarticletitle \undefined \def \showarticletitle #1{#1}   \fi
\ifx \showURL      \undefined \def \showURL       {\relax}        \fi
% invisible to TeX
\providecommand\bibfield[2]{#2}
\providecommand\bibinfo[2]{#2}
\providecommand\natexlab[1]{#1}
\providecommand\showeprint[2][]{arXiv:#2}

\bibitem[\protect\citeauthoryear{{AI}}{{AI}}{2019}]%
        {spacy}
\bibfield{author}{\bibinfo{person}{Explosion {AI}}.}
  \bibinfo{year}{2019}\natexlab{}.
\newblock \bibinfo{title}{{spaCy}: Industrial-Strength Natural Language
  Processing}.
\newblock
\newblock
\urldef\tempurl%
\url{https://spacy.io/}
\showURL{%
\tempurl}


\bibitem[\protect\citeauthoryear{Arora and Nyberg}{Arora and Nyberg}{2009}]%
        {arora2009interactive}
\bibfield{author}{\bibinfo{person}{Shilpa Arora} {and} \bibinfo{person}{Eric
  Nyberg}.} \bibinfo{year}{2009}\natexlab{}.
\newblock \showarticletitle{Interactive annotation learning with indirect
  feature voting}. In \bibinfo{booktitle}{\emph{Proc. NAACL-HLT Student
  Research Workshop and Doctoral Consortium}}.
\newblock


\bibitem[\protect\citeauthoryear{Chen}{Chen}{1976}]%
        {DBLP:journals/tods/Chen76}
\bibfield{author}{\bibinfo{person}{Peter~P. Chen}.}
  \bibinfo{year}{1976}\natexlab{}.
\newblock \showarticletitle{The Entity-Relationship Model - Toward a Unified
  View of Data}.
\newblock \bibinfo{journal}{\emph{{ACM} Trans. Database Syst.}}
  (\bibinfo{year}{1976}).
\newblock


\bibitem[\protect\citeauthoryear{et~al.}{et~al.}{2007a}]%
        {allen2007plow}
\bibfield{author}{\bibinfo{person}{Allen et al.}}
  \bibinfo{year}{2007}\natexlab{a}.
\newblock \showarticletitle{Plow: A collaborative task learning agent}. In
  \bibinfo{booktitle}{\emph{AAAI}}, Vol.~\bibinfo{volume}{7}.
  \bibinfo{pages}{1514--1519}.
\newblock


\bibitem[\protect\citeauthoryear{et~al.}{et~al.}{2017a}]%
        {bach2017learning}
\bibfield{author}{\bibinfo{person}{Bach et al.}}
  \bibinfo{year}{2017}\natexlab{a}.
\newblock \showarticletitle{Learning the structure of generative models without
  labeled data}. In \bibinfo{booktitle}{\emph{Proc. ICML}}.
\newblock


\bibitem[\protect\citeauthoryear{et~al.}{et~al.}{2009a}]%
        {druck2009active}
\bibfield{author}{\bibinfo{person}{Druck et al.}}
  \bibinfo{year}{2009}\natexlab{a}.
\newblock \showarticletitle{Active learning by labeling features}. In
  \bibinfo{booktitle}{\emph{Proc. EMNLP}}.
\newblock


\bibitem[\protect\citeauthoryear{et~al.}{et~al.}{2019a}]%
        {devlin2019bert}
\bibfield{author}{\bibinfo{person}{Devlin et al.}}
  \bibinfo{year}{2019}\natexlab{a}.
\newblock \showarticletitle{{BERT:} Pre-training of Deep Bidirectional
  Transformers for Language Understanding}. In \bibinfo{booktitle}{\emph{Proc.
  {NAACL-HLT}}}.
\newblock


\bibitem[\protect\citeauthoryear{et~al.}{et~al.}{2020a}]%
        {fu2020fast}
\bibfield{author}{\bibinfo{person}{Fu et al.}}
  \bibinfo{year}{2020}\natexlab{a}.
\newblock \bibinfo{title}{Fast and Three-rious: Speeding Up Weak Supervision
  with Triplet Methods}.
\newblock
\newblock
\showeprint[arxiv]{2002.11955}


\bibitem[\protect\citeauthoryear{et~al.}{et~al.}{2015a}]%
        {heer2015predictive}
\bibfield{author}{\bibinfo{person}{Heer et al.}}
  \bibinfo{year}{2015}\natexlab{a}.
\newblock \showarticletitle{Predictive Interaction for Data Transformation.}.
  In \bibinfo{booktitle}{\emph{Proc. CIDR}}.
\newblock


\bibitem[\protect\citeauthoryear{et~al.}{et~al.}{2017b}]%
        {DBLP:conf/ijcnn/HaldekarGO17}
\bibfield{author}{\bibinfo{person}{Haldekar et al.}}
  \bibinfo{year}{2017}\natexlab{b}.
\newblock \showarticletitle{Identifying spatial relations in images using
  convolutional neural networks}. In \bibinfo{booktitle}{\emph{{IJCNN}}}.
\newblock


\bibitem[\protect\citeauthoryear{et~al.}{et~al.}{2017c}]%
        {hanafi2017seer}
\bibfield{author}{\bibinfo{person}{Hanafi et al.}}
  \bibinfo{year}{2017}\natexlab{c}.
\newblock \showarticletitle{SEER: Auto-Generating Information Extraction Rules
  from User-Specified Examples}. In \bibinfo{booktitle}{\emph{Proc. CHI}}.
\newblock


\bibitem[\protect\citeauthoryear{et~al.}{et~al.}{2018a}]%
        {hancock2018training}
\bibfield{author}{\bibinfo{person}{Hancock et al.}}
  \bibinfo{year}{2018}\natexlab{a}.
\newblock \showarticletitle{Training classifiers with natural language
  explanations}. In \bibinfo{booktitle}{\emph{Proc. ACL}}.
\newblock


\bibitem[\protect\citeauthoryear{et~al.}{et~al.}{2019b}]%
        {jin2019clx}
\bibfield{author}{\bibinfo{person}{Jin et al.}}
  \bibinfo{year}{2019}\natexlab{b}.
\newblock \showarticletitle{{CLX:} Towards verifiable {PBE} data
  transformation}. In \bibinfo{booktitle}{\emph{Proc. EDBT}}.
\newblock


\bibitem[\protect\citeauthoryear{et~al.}{et~al.}{2011}]%
        {karger2011iterative}
\bibfield{author}{\bibinfo{person}{Karger et al.}}
  \bibinfo{year}{2011}\natexlab{}.
\newblock \showarticletitle{Iterative learning for reliable crowdsourcing
  systems}. In \bibinfo{booktitle}{\emph{NIPS}}.
\newblock


\bibitem[\protect\citeauthoryear{et~al.}{et~al.}{2003}]%
        {lau2003programming}
\bibfield{author}{\bibinfo{person}{Lau et al.}}
  \bibinfo{year}{2003}\natexlab{}.
\newblock \showarticletitle{Programming by demonstration using version space
  algebra}.
\newblock \bibinfo{journal}{\emph{Machine Learning}} (\bibinfo{year}{2003}).
\newblock


\bibitem[\protect\citeauthoryear{et~al.}{et~al.}{2008}]%
        {leshed2008coscripter}
\bibfield{author}{\bibinfo{person}{Leshed et al.}}
  \bibinfo{year}{2008}\natexlab{}.
\newblock \showarticletitle{CoScripter: automating \& sharing how-to knowledge
  in the enterprise}. In \bibinfo{booktitle}{\emph{Proc. CHI}}.
\newblock


\bibitem[\protect\citeauthoryear{et~al.}{et~al.}{2009b}]%
        {liang2009learning}
\bibfield{author}{\bibinfo{person}{Liang et al.}}
  \bibinfo{year}{2009}\natexlab{b}.
\newblock \showarticletitle{Learning from measurements in exponential
  families}. In \bibinfo{booktitle}{\emph{Proc. ICML}}.
\newblock


\bibitem[\protect\citeauthoryear{et~al.}{et~al.}{2017d}]%
        {li2017sugilite}
\bibfield{author}{\bibinfo{person}{Li et al.}}
  \bibinfo{year}{2017}\natexlab{d}.
\newblock \showarticletitle{SUGILITE: creating multimodal smartphone automation
  by demonstration}. In \bibinfo{booktitle}{\emph{Proc. CHI}}.
\newblock


\bibitem[\protect\citeauthoryear{et~al.}{et~al.}{2009c}]%
        {mintz2009distant}
\bibfield{author}{\bibinfo{person}{Mintz et al.}}
  \bibinfo{year}{2009}\natexlab{c}.
\newblock \showarticletitle{Distant supervision for relation extraction without
  labeled data}. In \bibinfo{booktitle}{\emph{Proc. ACL}}.
\newblock


\bibitem[\protect\citeauthoryear{et~al.}{et~al.}{2013}]%
        {menon2013machine}
\bibfield{author}{\bibinfo{person}{Menon et al.}}
  \bibinfo{year}{2013}\natexlab{}.
\newblock \showarticletitle{A machine learning framework for programming by
  example}. In \bibinfo{booktitle}{\emph{Proc. ICML}}.
\newblock


\bibitem[\protect\citeauthoryear{et~al.}{et~al.}{2018b}]%
        {peters2018elmo}
\bibfield{author}{\bibinfo{person}{Peters et al.}}
  \bibinfo{year}{2018}\natexlab{b}.
\newblock \showarticletitle{Deep Contextualized Word Representations}. In
  \bibinfo{booktitle}{\emph{Proc. {NAACL-HLT}}}.
\newblock


\bibitem[\protect\citeauthoryear{et~al.}{et~al.}{2005}]%
        {raghavan2005interactive}
\bibfield{author}{\bibinfo{person}{Raghavan et al.}}
  \bibinfo{year}{2005}\natexlab{}.
\newblock \showarticletitle{InterActive Feature Selection}. In
  \bibinfo{booktitle}{\emph{Proc. IJCAI}}.
\newblock


\bibitem[\protect\citeauthoryear{et~al.}{et~al.}{2016}]%
        {ratner2016data}
\bibfield{author}{\bibinfo{person}{Ratner et al.}}
  \bibinfo{year}{2016}\natexlab{}.
\newblock \showarticletitle{Data programming: Creating large training sets,
  quickly}. In \bibinfo{booktitle}{\emph{NIPS}}. \bibinfo{pages}{3567--3575}.
\newblock


\bibitem[\protect\citeauthoryear{et~al.}{et~al.}{2017e}]%
        {ratner2017snorkel}
\bibfield{author}{\bibinfo{person}{Ratner et al.}}
  \bibinfo{year}{2017}\natexlab{e}.
\newblock \showarticletitle{Snorkel: Rapid training data creation with weak
  supervision}.
\newblock \bibinfo{journal}{\emph{Proc. VLDB}} (\bibinfo{year}{2017}).
\newblock


\bibitem[\protect\citeauthoryear{et~al.}{et~al.}{2019c}]%
        {ratner2019training}
\bibfield{author}{\bibinfo{person}{Ratner et al.}}
  \bibinfo{year}{2019}\natexlab{c}.
\newblock \showarticletitle{Training complex models with multi-task weak
  supervision}. In \bibinfo{booktitle}{\emph{Proc. AAAI}}.
\newblock


\bibitem[\protect\citeauthoryear{et~al.}{et~al.}{2015b}]%
        {sculley2015hidden}
\bibfield{author}{\bibinfo{person}{Sculley et al.}}
  \bibinfo{year}{2015}\natexlab{b}.
\newblock \showarticletitle{Hidden technical debt in machine learning systems}.
  In \bibinfo{booktitle}{\emph{NIPS}}.
\newblock


\bibitem[\protect\citeauthoryear{et~al.}{et~al.}{2017f}]%
        {srivastava2017joint}
\bibfield{author}{\bibinfo{person}{Srivastava et al.}}
  \bibinfo{year}{2017}\natexlab{f}.
\newblock \showarticletitle{Joint concept learning and semantic parsing from
  natural language explanations}. In \bibinfo{booktitle}{\emph{Proc. EMNLP}}.
\newblock


\bibitem[\protect\citeauthoryear{et~al.}{et~al.}{2020b}]%
        {safranchik2020weakly}
\bibfield{author}{\bibinfo{person}{Safranchik et al.}}
  \bibinfo{year}{2020}\natexlab{b}.
\newblock \showarticletitle{Weakly Supervised Sequence Tagging from Noisy
  Rules}. In \bibinfo{booktitle}{\emph{Proc. AAAI}}, Vol.~\bibinfo{volume}{18}.
  \bibinfo{pages}{1--67}.
\newblock


\bibitem[\protect\citeauthoryear{et~al.}{et~al.}{2006}]%
        {von2006peekaboom}
\bibfield{author}{\bibinfo{person}{Von~Ahn et al.}}
  \bibinfo{year}{2006}\natexlab{}.
\newblock \showarticletitle{Peekaboom: a game for locating objects in images}.
  In \bibinfo{booktitle}{\emph{Proc. CHI}}.
\newblock


\bibitem[\protect\citeauthoryear{et~al.}{et~al.}{2007b}]%
        {zaidan2007using}
\bibfield{author}{\bibinfo{person}{Zaidan et al.}}
  \bibinfo{year}{2007}\natexlab{b}.
\newblock \showarticletitle{Using {``}Annotator Rationales{''} to Improve
  Machine Learning for Text Categorization}. In \bibinfo{booktitle}{\emph{Proc.
  NAACL-HLT}}.
\newblock


\bibitem[\protect\citeauthoryear{Gulwani}{Gulwani}{2011}]%
        {gulwani2011asp}
\bibfield{author}{\bibinfo{person}{Sumit Gulwani}.}
  \bibinfo{year}{2011}\natexlab{}.
\newblock \showarticletitle{Automating String Processing in Spreadsheets Using
  Input-output Examples}. In \bibinfo{booktitle}{\emph{Proc. POPL}}.
\newblock


\bibitem[\protect\citeauthoryear{Gupta and Manning}{Gupta and Manning}{2014}]%
        {gupta2014improved}
\bibfield{author}{\bibinfo{person}{Sonal Gupta} {and}
  \bibinfo{person}{Christopher Manning}.} \bibinfo{year}{2014}\natexlab{}.
\newblock \showarticletitle{Improved pattern learning for bootstrapped entity
  extraction}. In \bibinfo{booktitle}{\emph{Proc. CoNNL}}.
\newblock


\bibitem[\protect\citeauthoryear{Halbert}{Halbert}{1993}]%
        {halbert1993smallstar}
\bibfield{author}{\bibinfo{person}{Daniel~C. Halbert}.}
  \bibinfo{year}{1993}\natexlab{}.
\newblock \bibinfo{booktitle}{\emph{SmallStar: Programming by Demonstration in
  the Desktop Metaphor}}.
\newblock \bibinfo{publisher}{MIT Press}.
\newblock


\bibitem[\protect\citeauthoryear{Lacroix and Pirotte}{Lacroix and
  Pirotte}{1977}]%
        {DBLP:conf/vldb/LacroixP77}
\bibfield{author}{\bibinfo{person}{Michel Lacroix} {and} \bibinfo{person}{Alain
  Pirotte}.} \bibinfo{year}{1977}\natexlab{}.
\newblock \showarticletitle{Domain-Oriented Relational Languages}. In
  \bibinfo{booktitle}{\emph{Proc. VLDB}}.
\newblock


\bibitem[\protect\citeauthoryear{Lieberman}{Lieberman}{2001}]%
        {lieberman2001your}
\bibfield{author}{\bibinfo{person}{H. Lieberman}.}
  \bibinfo{year}{2001}\natexlab{}.
\newblock \bibinfo{booktitle}{\emph{Your Wish is My Command: Programming by
  Example}}.
\newblock \bibinfo{publisher}{Morgan Kaufmann Publishers}.
\newblock


\bibitem[\protect\citeauthoryear{McDaniel and Myers}{McDaniel and
  Myers}{1999}]%
        {mcdaniel1999getting}
\bibfield{author}{\bibinfo{person}{Richard McDaniel} {and}
  \bibinfo{person}{Brad Myers}.} \bibinfo{year}{1999}\natexlab{}.
\newblock \showarticletitle{Getting more out of programming-by-demonstration}.
  In \bibinfo{booktitle}{\emph{Proc. CHI}}.
\newblock


\bibitem[\protect\citeauthoryear{Myers}{Myers}{1998}]%
        {myers1998topaz}
\bibfield{author}{\bibinfo{person}{Brad~A Myers}.}
  \bibinfo{year}{1998}\natexlab{}.
\newblock \showarticletitle{Scripting graphical applications by demonstration}.
  In \bibinfo{booktitle}{\emph{Proc. CHI}}.
\newblock


\bibitem[\protect\citeauthoryear{Rennie and Lang}{Rennie and Lang}{2008}]%
        {rennie200820}
\bibfield{author}{\bibinfo{person}{Jason Rennie} {and} \bibinfo{person}{Ken
  Lang}.} \bibinfo{year}{2008}\natexlab{}.
\newblock \bibinfo{title}{The 20 Newsgroups data set}.
\newblock
\newblock


\bibitem[\protect\citeauthoryear{Varma and R{\'e}}{Varma and R{\'e}}{2018}]%
        {varma2018snuba}
\bibfield{author}{\bibinfo{person}{Paroma Varma} {and}
  \bibinfo{person}{Christopher R{\'e}}.} \bibinfo{year}{2018}\natexlab{}.
\newblock \bibinfo{title}{Snuba: automating weak supervision to label training
  data}.
\newblock
\newblock


\bibitem[\protect\citeauthoryear{Waldinger and Lee}{Waldinger and Lee}{1969}]%
        {waldinger1969prow}
\bibfield{author}{\bibinfo{person}{Richard~J. Waldinger} {and}
  \bibinfo{person}{Richard C.~T. Lee}.} \bibinfo{year}{1969}\natexlab{}.
\newblock \showarticletitle{PROW: A Step Toward Automatic Program Writing}. In
  \bibinfo{booktitle}{\emph{Proc. IJCAI}}.
\newblock


\bibitem[\protect\citeauthoryear{Zaidan and Eisner}{Zaidan and Eisner}{2008}]%
        {zaidan2008modeling}
\bibfield{author}{\bibinfo{person}{Omar Zaidan} {and} \bibinfo{person}{Jason
  Eisner}.} \bibinfo{year}{2008}\natexlab{}.
\newblock \showarticletitle{Modeling annotators: A generative approach to
  learning from annotator rationales}. In \bibinfo{booktitle}{\emph{Proc.
  EMNLP}}.
\newblock


\end{thebibliography}
